\documentclass{article}
\usepackage{amsmath}
\usepackage{amssymb}
\usepackage{amsthm}
\usepackage{algorithm}
\usepackage{algpseudocode}
\usepackage{geometry}
\usepackage{xcolor}
\usepackage[utf8]{inputenc}
\usepackage[T1]{fontenc}
\usepackage{hyperref}
\usepackage{cite}
\usepackage{booktabs}

\hypersetup{
    colorlinks=true,
    linkcolor=blue,
    filecolor=magenta,      
    urlcolor=cyan,
    citecolor=blue,
}

\newtheorem{theorem}{Theorem}
\newtheorem{lemma}[theorem]{Lemma}
\newtheorem{proposition}[theorem]{Proposition}

\newtheorem{remark}{Remark}
\newtheorem{definition}{Definition}

\newcommand{\E}{\mathbb{E}}
\newcommand{\grad}{\nabla_\theta}
\newcommand{\gradz}{\nabla_z}
\newcommand{\piStudent}{\pi_\theta}
\newcommand{\piRef}{\pi_{\text{ref}}}

\newcommand{\vocab}{\mathcal{V}}
\newcommand{\KL}{D_{\text{KL}}}

\title{\textbf{A Note on Hybrid Online Reinforcement and Imitation Learning for LLMs: Formulations and Algorithms}\thanks{First version: December 15, 2025}}
\author{Yingru Li, Ziniu Li, Jiacai Liu}
\date{}

\begin{document}

\maketitle

\begin{abstract}
We present a unified framework for Large Language Model (LLM) fine-tuning that integrates \textbf{Imitation Learning} and \textbf{Reinforcement Learning}. By analyzing the gradient of a composite objective combining trajectory-level KL divergence with task rewards, we derive a natural decomposition into two components: (1) an analytically computable \textbf{Dense Gradient} for token-level imitation, and (2) a Monte Carlo estimated \textbf{Sparse Gradient} for long-horizon reward optimization. The Dense Gradient admits a closed-form logit-level formula, enabling efficient GPU implementation.
\end{abstract}

\section{Introduction}

Knowledge Distillation (KD) \cite{hinton2015distilling} and Reinforcement Learning (RL) are two fundamental approaches to training Large Language Models. Standard KD minimizes divergence on static offline data via teacher forcing, leading to \textbf{train-inference distribution mismatch} (also known as exposure bias or covariate shift) \cite{bengio2015scheduled}: during training, the student conditions on ground-truth history, while during inference it conditions on its own generations. This mismatch compounds over long sequences, causing errors to accumulate---a phenomenon well-studied in online imitation learning \cite{ross2011dagger, rajaraman2021value}.

The DAgger algorithm \cite{ross2011dagger} addresses distribution mismatch by iteratively collecting data under the learner's own policy while querying the expert for labels. Recent theoretical work \cite{rajaraman2021value} establishes that interactive access to an expert provides provable statistical advantages over passive behavior cloning: under $\mu$-recoverability assumptions, interactive learners achieve suboptimality $\widetilde{O}(\mu |S| H / N)$, while non-interactive learners suffer $\Omega(|S| H^2 / N)$---a quadratic-to-linear improvement in horizon dependence.

RL addresses the distribution mismatch by training on self-generated data, but introduces new challenges: high variance from sparse, trajectory-level rewards and risks of reward hacking. Recent work on \textit{on-policy distillation} \cite{agarwal2024onpolicydistillationlanguagemodels, gu2024minillm} attempts to bridge these approaches. In this work, we analyze the gradient structure of a unified objective that combines on-policy KL minimization with reward maximization, revealing a natural decomposition that enables efficient implementation.

\paragraph{Contributions.}
\begin{enumerate}
    \item We derive the exact gradient of trajectory-level KL + reward objectives, proving it decomposes into \textbf{Dense} (analytic) and \textbf{Sparse} (sampled) terms (Theorem~\ref{thm:main}).
    \item We provide an efficient logit-level gradient formula amenable to GPU implementation (Proposition~\ref{prop:logit_grad}).
    \item We establish mathematical equivalence to KL-regularized RLHF while clarifying interpretational differences (Section~\ref{sec:rlhf}).
    \item We discuss training curriculum implications for the reward weight $\lambda$ (Section~\ref{sec:curriculum}).
\end{enumerate}

\section{Problem Formulation}

\subsection{Setup and Notation}

\begin{definition}[Autoregressive Policy]
An LLM policy $\pi$ generates response $y = (y_1, \ldots, y_T)$ given prompt $x$ with probability:
\begin{equation}
    \pi(y|x) = \prod_{t=1}^{T} \pi(y_t | x, y_{<t})
\end{equation}
where $y_{<t} = (y_1, \ldots, y_{t-1})$ and each $\pi(\cdot|x, y_{<t})$ is a distribution over vocabulary $\vocab$.
\end{definition}

\begin{definition}[Instantaneous Divergence Cost]
At step $t$, the log-ratio between student $\piStudent$ and reference $\piRef$ is:
\begin{equation}
    c_t = \log \piStudent(y_t | x, y_{<t}) - \log \piRef(y_t | x, y_{<t})
\end{equation}
\end{definition}

\subsection{The Hybrid Objective}

We minimize a cost combining distribution matching with reward maximization:
\begin{equation} \label{eq:objective}
    \mathcal{J}(\theta) = \E_{x \sim \mathcal{D}} \left[ \KL\big(\piStudent(\cdot|x) \,\|\, \piRef(\cdot|x)\big) - \lambda \, \E_{y \sim \piStudent} [r(x, y)] \right]
\end{equation}
where $\lambda \geq 0$ controls the trade-off and $r(x,y)$ is a black-box reward.

Using the autoregressive factorization, the trajectory KL decomposes as:
\begin{equation}
    \KL(\piStudent(\cdot|x) \| \piRef(\cdot|x)) = \E_{y \sim \piStudent}\left[\sum_{t=1}^T c_t\right]
\end{equation}

Thus the objective becomes:
\begin{equation} \label{eq:obj_expanded}
    \mathcal{J}(\theta) = \E_{x, y \sim \piStudent} \left[ \sum_{t=1}^{T} c_t - \lambda r(x, y) \right]
\end{equation}

\begin{remark}[Connection to Online Imitation Learning]
Our framework can be viewed as a continuous relaxation of DAgger \cite{ross2011dagger}. While DAgger iteratively collects data under the learner's distribution and trains via supervised learning, we directly optimize a trajectory-level objective under the learner's distribution. The Dense Term provides the imitation signal (analogous to DAgger's supervised loss), while on-policy sampling addresses the distribution mismatch.
\end{remark}

\section{Main Results} \label{sec:main}

We now state our main theoretical results. All proofs are deferred to Appendix~\ref{app:proofs}.

\begin{definition}[Future Return]
The future return from step $t+1$ is:
\begin{equation}
    G_{t+1} = \sum_{k=t+1}^{T} c_k - \lambda r(x, y)
\end{equation}
This captures future divergence costs plus the (negative) terminal reward.
\end{definition}

\begin{theorem}[Gradient Decomposition] \label{thm:main}
The gradient of objective \eqref{eq:objective} at step $t$ decomposes as:
\begin{equation} \label{eq:decomposition}
    \boxed{
    \grad \mathcal{J}_t(\theta) = \underbrace{\E \left[ \grad \log \piStudent(y_t|x, y_{<t}) \cdot c_t \right]}_{\textbf{Dense Term}} + \underbrace{\E \left[ \grad \log \piStudent(y_t|x, y_{<t}) \cdot G_{t+1} \right]}_{\textbf{Sparse Term}}
    }
\end{equation}
where the expectation is over $x \sim \mathcal{D}$ and $y \sim \piStudent(\cdot|x)$.
\end{theorem}

The proof relies on two key lemmas:

\begin{lemma}[Vanishing Score Function] \label{lem:vanishing}
For any context $(x, y_{<t})$:
\begin{equation}
    \E_{y_t \sim \piStudent(\cdot|x, y_{<t})} \left[\grad \log \piStudent(y_t|x, y_{<t})\right] = 0
\end{equation}
\end{lemma}

\begin{lemma}[Causality] \label{lem:causality}
For $k < t$, past costs do not contribute to the gradient at step $t$:
\begin{equation}
    \E_{y}\left[c_k \cdot \grad \log \piStudent(y_t|x, y_{<t})\right] = 0
\end{equation}
\end{lemma}

The Dense Term has a remarkable closed-form property:

\begin{proposition}[Dense Term Equals Token-Level KL Gradient] \label{prop:dense_kl}
\begin{equation}
    \E_{y_t \sim \piStudent} \left[ \grad \log \piStudent(y_t|x, y_{<t}) \cdot c_t \right] = \grad \KL\big(\piStudent(\cdot|x, y_{<t}) \,\|\, \piRef(\cdot|x, y_{<t})\big)
\end{equation}
\end{proposition}

This means the Dense Term can be computed \textbf{analytically} by summing over the finite vocabulary $\vocab$, without Monte Carlo sampling.

\begin{proposition}[Logit-Level Gradient] \label{prop:logit_grad}
Let $p = \mathrm{softmax}(z)$ be the student distribution and $q$ the teacher distribution. The gradient of KL with respect to logits $z$ is:
\begin{equation} \label{eq:logit_grad}
    \boxed{
    \gradz \KL(p\|q) = p \odot \left(\log p - \log q - \KL(p\|q) \cdot \mathbf{1}\right)
    }
\end{equation}
where $\odot$ denotes element-wise multiplication.
\end{proposition}

This formula enables efficient GPU implementation for both full-vocabulary and Top-$K$ computation.

\subsection{Interpretation of Terms}

\begin{center}
\begin{tabular}{@{}lll@{}}
\toprule
\textbf{Term} & \textbf{Dense (Imitation)} & \textbf{Sparse (RL)} \\
\midrule
Signal source & Current token distribution & Future trajectory + reward \\
Computation & Analytic (sum over $\vocab$) & Monte Carlo sampling \\
Variance & None (deterministic given context) & Scales with horizon $T$ \\
Contains reward? & No & Yes \\
\bottomrule
\end{tabular}
\end{center}

\subsection{Discounted Future Return}

To control the variance-bias trade-off, we introduce discount factor $\gamma \in [0,1]$:
\begin{equation}
    G_{t+1}^{(\gamma)} = \sum_{k=t+1}^{T} \gamma^{k-t} c_k - \lambda r(x, y)
\end{equation}

\textbf{Special cases:}
\begin{itemize}
    \item $\gamma = 0$: $G_{t+1}^{(0)} = -\lambda r(x,y)$ (only terminal reward, no future KL)
    \item $\gamma = 1$: $G_{t+1}^{(1)} = \sum_{k>t} c_k - \lambda r$ (full trajectory)
\end{itemize}

\begin{remark}[Discounting Convention]
The exponent $\gamma^{k-t}$ (rather than $\gamma^{k-t-1}$) means the immediate next step $c_{t+1}$ is discounted by $\gamma$ relative to the current cost $c_t$. This provides a natural interpolation: $\gamma = 0$ isolates each step from all future KL costs, while $\gamma = 1$ gives full trajectory credit assignment.
\end{remark}

\section{Algorithm}

\begin{algorithm}[H]
\caption{Hybrid Online Reinforcement \& Imitation Learning}
\label{alg:hybrid}
\begin{algorithmic}[1]
\State \textbf{Input:} Student $\piStudent$, Teacher $\piRef$, Reward $r$, Group size $K$, Discount $\gamma$, Weight $\lambda$

\For{each training iteration}
    \State Sample prompt $x \sim \mathcal{D}$
    \State Generate $K$ responses $\{y^{(i)}\}_{i=1}^K$ from $\piStudent(\cdot|x)$ \hfill \textit{// Group rollout}
    \State Initialize gradient accumulator $g \gets 0$
    
    \For{each response $i$, step $t$}
        \State $c_t^{(i)} \gets \log \piStudent(y_t^{(i)}|x, y_{<t}^{(i)}) - \log \piRef(y_t^{(i)}|x, y_{<t}^{(i)})$
    \EndFor
    
    \State Compute rewards: $R^{(i)} \gets r(x, y^{(i)})$
    
    \For{each response $i$, step $t$}
        \State $G_{t+1}^{(i)} \gets \sum_{k=t+1}^{T^{(i)}} \gamma^{k-t} c_k^{(i)} - \lambda R^{(i)}$
    \EndFor
    
    \For{each response $i$, step $t$}
        \State \textbf{Dense:} $g_{\text{dense}} \gets p \odot (\log p - \log q - \KL(p\|q) \cdot \mathbf{1})$ \hfill \textit{// Eq.~\eqref{eq:logit_grad}}
        \State \textbf{Sparse:} $g_{\text{sparse}} \gets \text{PolicyGradient}(\piStudent, y_t^{(i)}, G_{t+1}^{(i)})$
        \State $g \gets g + g_{\text{dense}} + g_{\text{sparse}}$ \hfill \textit{// Sum over all (i, t)}
    \EndFor
    \State $\theta \gets \theta - \eta \cdot g / K$ \hfill \textit{// Average over group}
\EndFor
\end{algorithmic}
\end{algorithm}

\begin{remark}[Compatibility with Group Rollout Infrastructure]
The algorithm is fully compatible with standard group rollout RL infrastructure. The group size $K$ corresponds to the number of responses sampled per prompt in existing implementations (e.g., GRPO, RLOO). The only additions are: (1) computing token-level log-ratios $c_t$ against the teacher, and (2) accumulating the Dense gradient alongside the standard policy gradient. Both operations integrate naturally into existing training pipelines.
\end{remark}

\section{Discussion}

\subsection{Training Curriculum for $\lambda$} \label{sec:curriculum}

The dual interpretation of our framework suggests a natural training curriculum:

\begin{itemize}
    \item \textbf{Early training} ($\lambda$ small, equivalently $\beta$ large): Prioritize imitation. Dense Term dominates, providing stable gradients toward teacher behavior (``cold start'').
    
    \item \textbf{Late training} ($\lambda$ large, equivalently $\beta$ small): Prioritize reward. Sparse Term becomes influential, enabling discovery of reward-maximizing behaviors beyond the teacher.
\end{itemize}

A simple schedule: $\lambda(t) = \lambda_0 (1 + \alpha t)$ where $t$ is training step.

In the limit $\lambda \to \infty$, the framework approaches pure RL, but the Dense Term provides ongoing regularization toward the expert.

\subsection{Relationship to KL-Regularized RLHF} \label{sec:rlhf}

Our objective \eqref{eq:objective} is mathematically equivalent to KL-regularized RLHF:
\begin{equation}
    \max_\theta \E_{y \sim \piStudent}[r(x,y)] - \beta \, \KL(\piStudent \| \piRef)
\end{equation}
with $\beta = 1/\lambda$. This follows by negating and rearranging: $\max \E[r] - \beta \KL \equiv \min \beta(\KL - \frac{1}{\beta}\E[r])$.

While mathematically identical, the frameworks differ in interpretation:

\begin{center}
\begin{tabular}{@{}lll@{}}
\toprule
\textbf{Aspect} & \textbf{KL-Regularized RLHF} & \textbf{Our Framework} \\
\midrule
Reference role & \textbf{Anchor} (starting point) & \textbf{Target} (expert teacher) \\
KL purpose & Regularization (prevent drift) & Imitation (attract to expert) \\
Typical $\piRef$ & SFT model (own initialization) & Stronger teacher model \\
\bottomrule
\end{tabular}
\end{center}

\subsection{Connection to Online Imitation Learning}

Our framework relates to the online imitation learning literature \cite{ross2011dagger, rajaraman2021value}:

\begin{center}
\begin{tabular}{@{}lll@{}}
\toprule
\textbf{Aspect} & \textbf{DAgger} & \textbf{Our Framework} \\
\midrule
Distribution & Learner's distribution & Learner's distribution \\
Expert signal & Discrete labels & Soft distribution (KL) \\
Optimization & Iterative supervised learning & End-to-end gradient \\
Reward integration & Separate & Unified via Sparse Term \\
\bottomrule
\end{tabular}
\end{center}

The theoretical results of \cite{rajaraman2021value} on the value of interaction suggest that our on-policy approach should provide statistical advantages over offline distillation, particularly for long-horizon tasks where distribution mismatch is severe.

\subsection{Relationship to DAPO}

Divergence-Augmented Policy Optimization (DAPO) \cite{wang2019dapo} also incorporates divergence terms into policy optimization. DAPO adds a Bregman divergence between the \emph{behavior policy} (which generated the training data) and the current policy to stabilize \emph{off-policy} learning---the divergence controls the degree of ``off-policyness'' to ensure safe policy updates when reusing old data.

\textbf{Common ground.} Both frameworks operate on divergence over state-action occupancy measures. In the autoregressive LLM setting, sequence-level KL is equivalent to KL over state-action occupancy: the ``state'' is the context $(x, y_{<t})$ and the ``action'' is the next token $y_t$. The trajectory distribution $\pi(y|x) = \prod_t \pi(y_t|x,y_{<t})$ induces a state-action occupancy, and our trajectory-level KL decomposes accordingly. DAPO similarly computes Bregman divergence over state-action distributions rather than action probabilities alone.

\textbf{Key differences.} Despite operating on the same mathematical objects, the frameworks differ in purpose:

\begin{center}
\begin{tabular}{@{}lll@{}}
\toprule
\textbf{Aspect} & \textbf{Our Framework} & \textbf{DAPO} \\
\midrule
Reference policy & Expert/teacher $\piRef$ & Behavior policy $\pi_b$ \\
Divergence purpose & Imitation (attract to expert) & Stability (bound off-policyness) \\
Data regime & On-policy & Off-policy \\
\bottomrule
\end{tabular}
\end{center}

Our KL term drives the policy \emph{toward} a reference expert, while DAPO's divergence term prevents the policy from drifting \emph{too far} from the data-generating behavior policy. Our on-policy setting eliminates the need for importance weights that DAPO requires for off-policy correction.

\subsection{Unification of Existing Methods}

Different choices of $\gamma$ and $\lambda$ recover existing approaches:

\begin{center}
\begin{tabular}{@{}lccl@{}}
\toprule
\textbf{Method} & $\gamma$ & $\lambda$ & \textbf{Active Terms} \\
\midrule
Standard SFT/KD & 0 & 0 & Dense only \\
On-policy distillation + reward & 0 & $>0$ & Dense + reward via Sparse \\
Full trajectory KL + reward & 1 & $>0$ & Dense + full Sparse \\
Pure RL (no KL) & --- & $>0$ & Sparse only \\
\bottomrule
\end{tabular}
\end{center}

\subsection{Extensions}

The framework naturally extends to multiple teachers and rewards:

\textbf{Multiple Teachers.} Given teachers $\{\pi_{\text{ref}}^{(m)}\}_{m=1}^M$ with weights $\alpha_m \geq 0$:
\begin{equation}
    \text{Dense}_t = \sum_{m=1}^{M} \alpha_m \grad \KL\big(\piStudent(\cdot|x, y_{<t}) \,\|\, \pi_{\text{ref}}^{(m)}(\cdot|x, y_{<t})\big)
\end{equation}

\textbf{Multiple Rewards.} Given rewards $\{r_n\}_{n=1}^N$ with weights $\lambda_n$:
\begin{equation}
    G_{t+1}^{(\gamma)} = \sum_{k=t+1}^{T} \gamma^{k-t} c_k - \sum_{n=1}^{N} \lambda_n r_n(x, y)
\end{equation}

\section{Conclusion}

We derived a principled gradient decomposition for hybrid imitation-reinforcement learning in LLMs:
\begin{enumerate}
    \item The gradient splits into \textbf{Dense} (analytic, zero sampling variance) and \textbf{Sparse} (Monte Carlo, captures long-horizon effects) terms.
    \item The logit-level formula \eqref{eq:logit_grad} enables efficient GPU implementation.
    \item Mathematical equivalence to RLHF ($\beta = 1/\lambda$) is established, connecting to the broader literature on KL-regularized policy optimization.
    \item The framework suggests natural training curricula via $\lambda$ scheduling, transitioning from imitation-focused to reward-focused learning.
\end{enumerate}

Future work includes empirical validation and analysis of variance-bias trade-offs for different $\gamma$ values.


\appendix

\section{Proofs} \label{app:proofs}

\subsection{Proof of Lemma~\ref{lem:vanishing} (Vanishing Score Function)}

\begin{proof}
For any distribution $\piStudent(\cdot|x, y_{<t})$ over finite vocabulary $\vocab$:
\begin{align}
    \E_{y_t \sim \piStudent}\left[\grad \log \piStudent(y_t|x, y_{<t})\right] 
    &= \sum_{v \in \vocab} \piStudent(v|x, y_{<t}) \cdot \frac{\grad \piStudent(v|x, y_{<t})}{\piStudent(v|x, y_{<t})} \\
    &= \sum_{v \in \vocab} \grad \piStudent(v|x, y_{<t}) \\
    &= \grad \sum_{v \in \vocab} \piStudent(v|x, y_{<t}) \\
    &= \grad(1) = 0
\end{align}
The key step uses that probabilities sum to 1, which is constant with respect to $\theta$.
\end{proof}

\subsection{Proof of Lemma~\ref{lem:causality} (Causality)}

\begin{proof}
For $k < t$, the cost $c_k$ depends only on tokens $y_1, \ldots, y_k$. We decompose the expectation using the tower property:
\begin{align}
    &\E_{y}\left[c_k \cdot \grad \log \piStudent(y_t|x, y_{<t})\right] \\
    &= \E_{y_{<t}}\left[c_k \cdot \E_{y_t | y_{<t}}\left[\grad \log \piStudent(y_t|x, y_{<t})\right]\right] \\
    &= \E_{y_{<t}}\left[c_k \cdot 0\right] \quad \text{(by Lemma~\ref{lem:vanishing})} \\
    &= 0
\end{align}
Since $c_k$ is measurable with respect to $y_{<t}$ for $k < t$, it factors out of the inner expectation.
\end{proof}

\subsection{Proof of Theorem~\ref{thm:main} (Gradient Decomposition)}

\begin{proof}
Define total cost $C(y) = \sum_{t=1}^T c_t - \lambda r(x,y)$.

\textbf{Step 1: Apply REINFORCE identity \cite{williams1992simple}.}

For any function $f(y)$:
\begin{equation}
    \grad \E_{y \sim \piStudent}[f(y)] = \E_y\left[\grad f(y) + f(y) \cdot \grad \log \piStudent(y|x)\right]
\end{equation}

Applying this to $C(y)$:
\begin{equation}
    \grad \E_{y}[C(y)] = \E_y\left[\grad C(y) + C(y) \cdot \grad \log \piStudent(y|x)\right]
\end{equation}

\textbf{Step 2: Show $\E_y[\grad C(y)] = 0$.}

Since $r$ and $\piRef$ don't depend on $\theta$:
\begin{equation}
    \grad C(y) = \sum_{t=1}^T \grad c_t = \sum_{t=1}^T \grad \log \piStudent(y_t|x, y_{<t})
\end{equation}

Taking expectation and using iterated conditioning:
\begin{equation}
    \E_y[\grad c_t] = \E_{y_{<t}}\left[\E_{y_t|y_{<t}}[\grad \log \piStudent(y_t|x, y_{<t})]\right] = \E_{y_{<t}}[0] = 0
\end{equation}
by Lemma~\ref{lem:vanishing}. Thus $\E_y[\grad C(y)] = 0$.

\textbf{Step 3: Expand $\grad \log \piStudent(y|x)$.}

By the chain rule on the autoregressive factorization:
\begin{equation}
    \grad \log \piStudent(y|x) = \sum_{k=1}^T \grad \log \piStudent(y_k|x, y_{<k})
\end{equation}

\textbf{Step 4: Apply causality to isolate per-step contributions.}

The gradient contribution at step $t$ is:
\begin{equation}
    \E_y\left[C(y) \cdot \grad \log \piStudent(y_t|x, y_{<t})\right]
\end{equation}

Decompose $C(y) = \sum_{k<t} c_k + c_t + G_{t+1}$ where $G_{t+1} = \sum_{k>t} c_k - \lambda r$.

By Lemma~\ref{lem:causality}, past costs ($k < t$) contribute zero:
\begin{equation}
    \E_y\left[\sum_{k<t} c_k \cdot \grad \log \piStudent(y_t|x, y_{<t})\right] = 0
\end{equation}

Thus:
\begin{equation}
    \grad \mathcal{J}_t = \E_y\left[(c_t + G_{t+1}) \cdot \grad \log \piStudent(y_t|x, y_{<t})\right]
\end{equation}

Linearity of expectation gives the Dense ($c_t$) + Sparse ($G_{t+1}$) decomposition.
\end{proof}

\subsection{Proof of Proposition~\ref{prop:dense_kl} (Dense Term = KL Gradient)}

\begin{proof}
\textbf{Left-hand side (Dense Term):}
\begin{align}
    \E_{y_t}[c_t \cdot \grad \log \piStudent(y_t)] 
    &= \sum_{v} \piStudent(v) \cdot \log\frac{\piStudent(v)}{\piRef(v)} \cdot \frac{\grad \piStudent(v)}{\piStudent(v)} \\
    &= \sum_{v} \grad \piStudent(v) \cdot \log\frac{\piStudent(v)}{\piRef(v)}
\end{align}

\textbf{Right-hand side (KL Gradient):}
\begin{align}
    \grad \KL(\piStudent \| \piRef) &= \grad \sum_v \piStudent(v) \log\frac{\piStudent(v)}{\piRef(v)} \\
    &= \sum_v \left[\grad \piStudent(v) \cdot \log\frac{\piStudent(v)}{\piRef(v)} + \piStudent(v) \cdot \frac{\grad \piStudent(v)}{\piStudent(v)}\right] \\
    &= \sum_v \grad \piStudent(v) \cdot \log\frac{\piStudent(v)}{\piRef(v)} + \sum_v \grad \piStudent(v) \\
    &= \sum_v \grad \piStudent(v) \cdot \log\frac{\piStudent(v)}{\piRef(v)} + \grad(1) \\
    &= \sum_v \grad \piStudent(v) \cdot \log\frac{\piStudent(v)}{\piRef(v)}
\end{align}

LHS = RHS. $\square$
\end{proof}

\subsection{Proof of Proposition~\ref{prop:logit_grad} (Logit-Level Gradient)}

\begin{proof}
Let $p = \text{softmax}(z)$ where $p_i = \frac{e^{z_i}}{\sum_j e^{z_j}}$.

\textbf{Step 1: Compute the softmax Jacobian.}

Taking the derivative:
\begin{equation}
    \frac{\partial p_i}{\partial z_j} = p_i(\delta_{ij} - p_j)
\end{equation}
where $\delta_{ij}$ is the Kronecker delta.

\textbf{Step 2: Apply chain rule for KL.}
\begin{equation}
    \frac{\partial \KL}{\partial z_j} = \sum_i \frac{\partial \KL}{\partial p_i} \cdot \frac{\partial p_i}{\partial z_j}
\end{equation}

Since $\KL = \sum_i p_i(\log p_i - \log q_i)$:
\begin{equation}
    \frac{\partial \KL}{\partial p_i} = \log p_i - \log q_i + 1
\end{equation}

\textbf{Step 3: Combine using the Jacobian.}
\begin{align}
    \frac{\partial \KL}{\partial z_j} &= \sum_i p_i(\delta_{ij} - p_j)(\log p_i - \log q_i + 1) \\
    &= p_j(\log p_j - \log q_j + 1) - p_j \sum_i p_i(\log p_i - \log q_i + 1) \\
    &= p_j(\log p_j - \log q_j + 1) - p_j(\KL + 1) \\
    &= p_j(\log p_j - \log q_j - \KL)
\end{align}

In vector form: $\gradz \KL = p \odot (\log p - \log q - \KL \cdot \mathbf{1})$.
\end{proof}

\section{Implementation Details} \label{app:implementation}

\subsection{Full Vocabulary Computation}

Using Equation~\eqref{eq:logit_grad}, the dense gradient can be computed as follows (pseudocode):

\begin{verbatim}
def dense_gradient_full(student_logits, teacher_logits):
    p = softmax(student_logits)
    q = softmax(teacher_logits)
    kl = sum(p * (log(p) - log(q)))
    grad_z = p * (log(p) - log(q) - kl)
    return grad_z
\end{verbatim}

Complexity: $O(|\vocab|)$ per position. The element-wise operations are amenable to GPU parallelization (e.g., via custom CUDA or Triton kernels).

\subsection{Top-K Approximation}

For computational efficiency when $|\vocab|$ is large (pseudocode):

\begin{verbatim}
def dense_gradient_topk(student_logits, teacher_logits, K=32):
    # Select top-K from STUDENT distribution
    top_k_idx = topk_indices(student_logits, K)
    
    p_full = softmax(student_logits)
    q_full = softmax(teacher_logits)
    
    # NO renormalization - use raw probabilities
    p = p_full[top_k_idx]
    q = q_full[top_k_idx]
    
    kl_approx = sum(p * (log(p) - log(q)))
    
    grad_z = zeros_like(student_logits)
    grad_z[top_k_idx] = p * (log(p) - log(q) - kl_approx)
    return grad_z
\end{verbatim}

\textbf{Critical implementation notes:}
\begin{enumerate}
    \item Top-K indices from \textbf{student} (targets tokens being considered)
    \item \textbf{No renormalization} (naturally scales gradient magnitude based on confidence)
    \item \textbf{Approximation:} Using \texttt{kl\_approx} (sum over top-K) instead of full $\KL$ introduces bias; this is acceptable when top-K captures most probability mass
\end{enumerate}

\subsection{Complexity Comparison}

\begin{center}
\begin{tabular}{@{}lcc@{}}
\toprule
\textbf{Method} & \textbf{Gradient Time} & \textbf{Gradient Memory} \\
\midrule
Full vocabulary & $O(|\vocab|)$ & $O(|\vocab|)$ \\
Top-$K$ & $O(K)$ & $O(K)$ \\
\bottomrule
\end{tabular}
\end{center}

Note: Both methods require $O(|\vocab|)$ for the initial softmax computation. The savings apply to the gradient tensor: Top-$K$ produces a sparse gradient with only $K$ non-zero entries. For $|\vocab| \approx 128{,}000$ and $K = 32$, this reduces gradient storage and downstream computation by $\sim$4000$\times$.

\end{document}